\documentclass[11pt]{article}
\usepackage{amssymb}

\usepackage[utf8]{inputenc}
\usepackage[T1]{fontenc}
\usepackage{graphicx}
\usepackage{booktabs}
\usepackage{adjustbox} 
\usepackage{array}
\usepackage{longtable}
\usepackage{subcaption}
\usepackage{ulem}
\usepackage[table]{xcolor}
\usepackage{makecell} 
\usepackage{siunitx} 
\usepackage{bbm}
\usepackage{algpseudocode}

\usepackage[T1]{fontenc}    % use 8-bit T1 fonts
\usepackage{hyperref}       % hyperlinks
\usepackage{url}            % simple URL typesetting
\usepackage{amsfonts}       % blackboard math symbols
\usepackage{amsmath}        
\usepackage{nicefrac}       
\usepackage{microtype}      % microtypography
\usepackage{xcolor}         % colors
\usepackage{caption}
\usepackage{subcaption}
\usepackage{placeins}
\usepackage{lscape}
\usepackage{array}
\usepackage{float} 
\usepackage{algorithm}
\usepackage{xcolor}

\usepackage{graphicx, makecell, booktabs}

\usepackage{multirow}
\usepackage[final]{acl}

\usepackage{times}
\usepackage{latexsym}
\usepackage{amsmath}
\usepackage{color,colortbl}
\usepackage{comment}
\usepackage[T1]{fontenc}
\usepackage[utf8]{inputenc}

\usepackage{microtype}

\usepackage{inconsolata}

\usepackage{graphicx}
\usepackage{xspace}
\definecolor{Gray}{gray}{0.85}

\title{The Token Tax: Systematic Bias in Multilingual Tokenization}

\author{
 \textbf{Jessica M.~Lundin\textsuperscript{1}},
 \textbf{Ada Zhang\textsuperscript{2}},
 \textbf{Nihal Karim\textsuperscript{2}},
 \textbf{Hamza Louzan\textsuperscript{2}},
\\
 \textbf{Victor Wei\textsuperscript{2}},
 \textbf{David Ifeoluwa Adelani\textsuperscript{3}},
 \textbf{Cody Carroll\textsuperscript{2,4}}
\\
\\
 \textsuperscript{1}Institute for Disease Modeling, Gates Foundation,\\
 \textsuperscript{2}Data Institute, University of San Francisco,\\
 \textsuperscript{3}Mila - Quebec AI Institute, McGill University, and Canada CIFAR AI Chair,\\
 \textsuperscript{4}Department of Mathematics and Statistics, University of San Francisco
\\
 \small{
   \textbf{Correspondence:} \href{mailto:jessica.lundin@gatesfoundation.org}{jessica.lundin@gatesfoundation.org}
 }
}

\begin{document}
\maketitle

\begin{abstract}
Tokenization inefficiency is associated with structural disadvantages for morphologically complex, low-resource languages, inflating compute resources and reducing accuracy. We evaluate 11 large language models (LLMs) on AfriMMLU (5 subjects; 16 African languages); for the 10 models whose fertility we could compute, higher token fertility is associated with lower accuracy: all 50 model--subject slopes are negative, 24 are significant at $p<0.05$, and 12 remain significant after false-discovery-rate correction.  We further find that reasoning models (e.g., DeepSeek R1, o1) have the highest average accuracy and the smallest English–African gaps among the models evaluated. In terms of economics, a doubling in tokens results in quadrupled training cost and time, underscoring the "token tax" faced by many languages. These results motivate morphologically aware tokenization, fair pricing, and multilingual benchmarks for equitable natural language processing (NLP).

\end{abstract}

\section{Introduction}

Tokenization serves as the foundational layer of modern NLP systems, yet it is associated with systematic inequalities that disproportionately affect morphologically complex and low-resource languages. Prior work decisively establishes tokenization as a source of computational and economic inequality~\citep{ahia-etal-2023-languages}, with quantified impacts ranging from inflated token counts to substantial BLEU point performance degradation \cite{petrov2023, ali-etal-2024-tokenizer}.  Because transformer attention scales quadratically with sequence length, even modest increases in token counts can meaningfully raise compute requirements and reduce effective context capacity \cite{keles2022computationalcomplexityselfattention}. As a result, morphologically complex languages with high fertility values suffer from compound disadvantages that are difficult to fully overcome within widely deployed transformer architectures \cite{sreedhar2023localbytefusionneural}. Recent work on adaptive and flexible tokenization strategies (e.g., MAGNET; FlexiTokens) suggests promising directions for mitigating these inefficiencies, but these approaches are not yet standard in large-scale multilingual LLM training and remain mostly unevaluated on low-resource African languages \cite{ahia2024magnet, owodunni2025flexitokens}.

These disparities particularly affect morphologically rich languages, where agglutinative and fusional morphology leads to systematic tokenization inefficiency. The technical disparities translate directly into economic exclusion through what we term the "token tax," prohibitive training and inference costs measured in dollars and tons of CO$_2$, and systematic underrepresentation in model capabilities that affects billions of speakers worldwide.

A reasonable cost to train a small-medium model or a large frontier model is easily \$1M (1 month) to \$100M ($\sim$3 months) with primarily English tokens (see Appendix C for derivation of estimates from publicly reported petaFLOP-day requirements and standard cloud compute pricing). If we instead train on a language with 2× or 5× more tokens for the same content, the transformer's quadratic $O(n^2)$ compute costs result in a 4× or 25× increase in energy consumption, dollar cost, training time, and CO$_2$ emissions relative to English. In this example, the cost becomes \$4-25M (4 months-2 years) or \$400M-2.5B (1-6 years), respectively.

Our contributions herein are as follows:
\begin{itemize}
\item We extend prior fertility and accuracy analysis to 10 models and 16 languages, finding a consistently negative association between fertility and MCQA accuracy that is statistically significant in about half of the model--subject regressions.
\item We conduct the first large-scale comparison of tokenization effects for reasoning vs. non-reasoning LLMs on AfriMMLU, showing that reasoning models narrow, but do not close, the English–African accuracy gap.
\item We quantify the economic impact of tokenization inefficiency, demonstrating how the "token tax" creates barriers to multilingual NLP development.
\item We release public datasets containing: (i) model results from AfriMMLU benchmark including reasoning models, (ii) comprehensive tokenization metrics.
\end{itemize}

\section{Related Work}

% Tokenization and Multilingual NLP
The impact of tokenization on multilingual model performance has received increasing attention. \citet{petrov2023} demonstrated that tokenizers disadvantage non-English languages, with text length exceeding 15 times for some language pairs. \citet{rust2021good} showed that tokenizer quality significantly impacts downstream performance, finding that morphologically complex languages suffer from both vocabulary underrepresentation and suboptimal segmentation strategies. \citet{ali-etal-2024-tokenizer} conducted extensive ablation experiments showing that tokenizer choice can impact downstream performance and increase training costs by 68\%. Their analysis is for European languages that have different fertility distributions than African languages, the focus of this work.

% \paragraph{Systematic Inequalities in NLP.}
\citet{joshi-etal-2020-state} categorized languages into six resource levels, showing that 88\% of world languages fall into the lowest category. These systemic inequalities correlate strongly with tokenization efficiency, suggesting that addressing tokenization could substantially improve language technology equity.

% African Language Evaluation
Recent benchmarks have begun addressing the evaluation gap for African languages \cite{afrimmlu2024, singh2025globalmmluunderstandingaddressing, alhanai2024bridginggapenhancingllm, adebara-etal-2025-evaluating, beyene2025mstebmassivelymultilingualevaluation, nllbteam2022languageleftbehindscaling}. While these benchmarks evaluate various aspects of multilingual performance, from culturally relevant knowledge questions to speech tasks, they primarily focus on documenting performance gaps without systematically analyzing the underlying tokenization disparities. Our work complements these efforts by explicitly connecting tokenization efficiency to performance degradation, showing that performance gaps are associated with tokenization inefficiency, although fertility is confounded with training-data availability and we do not separate the two.

% Economic and Computational Costs.
While the computational complexity of transformers is well-established \cite{vaswani2023attentionneed}, the economic implications of tokenization inefficiency remain underexplored. Our work bridges this gap by quantifying both performance degradation and financial costs, demonstrating how technical choices create economic barriers to equitable NLP development.

\section{Experimental Setup}
\label{sec:setup}

\subsection{Dataset and Languages}
AfriMMLU \cite{afrimmlu2024} comprises 9,000 multiple-choice questions across 5 subjects: elementary mathematics, global facts, high school geography, high school microeconomics, and international law. The benchmark covers 16 African languages spanning multiple language families including Niger-Congo (e.g., Swahili, Yoruba, Zulu), Afro-Asiatic (e.g., Amharic, Hausa), and Nilo-Saharan, providing substantial typological diversity. Each subject contains between 1,500 and 2,000 questions, professionally translated by native speakers with quality verification.

\subsection{Model Selection}
We evaluate 11 models spanning three categories to enable comparison, including general LLMs (Llama 3.1 405B, Gemini 1.5 Pro, Claude Sonnet 3.5, DeepSeek V3, GPT-4o, Qwen 2.5 32B, Phi-4) and multilingual-focused ones (Aya 23 35B, Pixtral 12B). We also include reasoning models (DeepSeek R1, OpenAI o1) to test whether enhanced reasoning capabilities mitigate tokenization-related performance decline. Because the tokenizer of o1 is not public, the fertility regressions cover the other 10 models.  If reasoning models show reduced gaps despite similar tokenization inefficiency, this would suggest that architectural advances can partially mitigate tokenizer limitations.  

\subsection{Metrics and Analysis}

Our primary metric is Fertility ($F = T~/~W$), the average number of tokens per word, quantifying tokenization efficiency. Lower fertility indicates more efficient tokenization for a given language.  Our analysis also includes zero-shot accuracy of MCQA tasks using the prompt template detailed in the Appendix.

Performance gaps are calculated as the difference between English and French accuracy and the mean accuracy across all 16 African languages for each model-subject combination.  English and French here provide baselines near optimal tokenization efficiency.  The "Random" baseline of 25\% represents the probability of guessing a correct response by chance for multiple choice questions with 4 options.

\subsection{Statistical Methodology}

The pipeline for analysis is as follows:
\begin{algorithm}
\caption{Fertility-Accuracy Analysis Pipeline}
\label{alg:pipeline}
\begin{algorithmic}[1]
\For{each language $\ell$ in AfriMMLU}
   \For{each model $m$}
       \State Tokenize all questions using $m$'s tokenizer
       \State Calculate fertility $F_{\ell,m} = \frac{\text{tokens}}{\text{words}}$
       \State Evaluate zero-shot accuracy on all subjects
       \State Store results for regression analysis
   \EndFor
\EndFor
\State Fit linear models: Accuracy $\sim$ Fertility for each model-subject pair
\State Calculate regression statistics: slopes, standard errors, t-values, p-values
\State Calculate Pearson correlation coefficients ($\rho$), $R^2$, and adjusted $R^2$
\State Apply Benjamini-Hochberg FDR correction for multiple comparisons
\end{algorithmic}
\end{algorithm}

We conduct linear regression analysis for each model-subject combination, treating fertility as the predictor and accuracy as the outcome variable. For each regression, we report the intercept, slope (effect of fertility on accuracy), standard error, $t$-statistic, and $p$-value. We calculate Pearson correlation coefficients ($\rho$) to quantify the strength and direction of the linear relationship, and coefficients of determination ($R^2$) to measure the proportion of variance in accuracy explained by fertility. Adjusted $R^2$ values correct for sample size and number of predictors. Statistical significance is assessed at $p < 0.05$ with Benjamini-Hochberg False Discovery Rate (FDR) correction to account for 
multiple comparisons across 50 model-subject pairs. We consider $R^2 \geq 0.25$ as large effects and $|\rho| \geq 0.50$ as strong correlations following standard conventions.

\section{Results}

\begin{table*}[!ht]
\centering

\resizebox{.95\textwidth}{!}{
\begin{tabular}{llccccccccc}
\toprule
\textbf{Subject} & \textbf{Model} & \textbf{Intercept} & \textbf{Slope} & \textbf{Std. Error} & \textbf{t-value} & \textbf{P-value} & \textbf{$\rho$} & \textbf{$R^2$} & \textbf{Adj. $R^2$} \\
\midrule
Elementary Math & Sonnet3.5 & 0.652 & -0.018 & 0.029 & -0.609 & 0.552 & -0.155 & 0.024 & -0.041 \\
 & Aya23 35B & 0.480 & -0.079 & 0.019 & -4.058 & \textbf{0.001*} & \textbf{-0.723} & \textbf{0.523} & \textbf{0.492} \\
 & DeepSeek R1 & 1.045 & -0.066 & 0.045 & -1.475 & 0.161 & -0.356 & 0.127 & 0.068 \\
 & DeepSeek V3 & 0.884 & -0.044 & 0.048 & -0.922 & 0.371 & -0.232 & 0.054 & -0.009 \\
 & Gemini 1.5 Pro & 0.907 & -0.045 & 0.042 & -1.078 & 0.298 & -0.268 & 0.072 & 0.010 \\
 & Llama3.1 405B & 0.836 & -0.054 & 0.045 & -1.195 & 0.251 & -0.295 & 0.087 & 0.026 \\
 & Phi4 & 0.773 & -0.125 & 0.034 & -3.641 & \textbf{0.002*} & \textbf{-0.685} & 0.469 & 0.434 \\
 & GPT-4o & 1.002 & -0.089 & 0.057 & -1.571 & 0.137 & -0.376 & 0.141 & 0.084 \\
 & Pixtral 12B & 0.417 & -0.024 & 0.014 & -1.717 & 0.106 & -0.405 & 0.164 & 0.109 \\
 & Qwen2.5 32B & 0.857 & -0.113 & 0.037 & -3.012 & \textbf{0.009*} & \textbf{-0.614} & 0.377 & 0.335 \\
\midrule
Global Facts & Sonnet3.5 & 0.508 & -0.011 & 0.027 & -0.390 & 0.702 & -0.100 & 0.011 & -0.056 \\
 & Aya23 35B & 0.335 & -0.044 & 0.023 & -1.930 & 0.073 & -0.446 & 0.199 & 0.146 \\
 & DeepSeek R1 & 0.574 & -0.002 & 0.038 & -0.061 & 0.952 & -0.016 & 0.000 & -0.066 \\
 & DeepSeek V3 & 0.619 & -0.045 & 0.034 & -1.308 & 0.211 & -0.320 & 0.102 & 0.042 \\
 & Gemini 1.5 Pro & 0.585 & -0.011 & 0.051 & -0.222 & 0.827 & -0.057 & 0.003 & -0.063 \\
 & Llama3.1 405B & 0.685 & -0.084 & 0.033 & -2.564 & \textbf{0.022} & \textbf{-0.552} & \textbf{0.305} & \textbf{0.258} \\
 & Phi4 & 0.408 & -0.015 & 0.017 & -0.845 & 0.411 & -0.213 & 0.045 & -0.018 \\
 & GPT-4o & 0.638 & -0.063 & 0.054 & -1.169 & 0.261 & -0.289 & 0.084 & 0.022 \\
 & Pixtral 12B & 0.428 & -0.038 & 0.018 & -2.068 & 0.056 & -0.471 & 0.222 & 0.170 \\
 & Qwen2.5 32B & 0.505 & -0.052 & 0.024 & -2.171 & \textbf{0.046} & -0.489 & 0.239 & 0.188 \\
\midrule
High School Geography & Sonnet3.5 & 0.781 & -0.080 & 0.045 & -1.779 & 0.096 & -0.417 & 0.174 & 0.119 \\
 & Aya23 35B & 0.475 & -0.097 & 0.038 & -2.512 & \textbf{0.024} & \textbf{-0.544} & 0.296 & 0.249 \\
 & DeepSeek R1 & 0.847 & -0.082 & 0.056 & -1.466 & 0.163 & -0.354 & 0.125 & 0.067 \\
 & DeepSeek V3 & 0.843 & -0.124 & 0.053 & -2.331 & \textbf{0.034} & \textbf{-0.516} & 0.266 & 0.217 \\
 & Gemini 1.5 Pro & 0.750 & -0.065 & 0.070 & -0.937 & 0.363 & -0.235 & 0.055 & -0.008 \\
 & Llama3.1 405B & 0.822 & -0.131 & 0.055 & -2.359 & \textbf{0.032} & \textbf{-0.520} & 0.271 & 0.222 \\
 & Phi4 & 0.808 & -0.162 & 0.048 & -3.343 & \textbf{0.004*} & \textbf{-0.653} & 0.427 & 0.389 \\
 & GPT-4o & 0.952 & -0.151 & 0.068 & -2.211 & \textbf{0.043} & -0.496 & 0.246 & 0.195 \\
 & Pixtral 12B & 0.688 & -0.121 & 0.035 & -3.414 & \textbf{0.004*} & \textbf{-0.661} & \textbf{0.437} & \textbf{0.400} \\
 & Qwen2.5 32B & 0.755 & -0.155 & 0.049 & -3.190 & \textbf{0.006*} & \textbf{-0.636} & 0.404 & 0.365 \\
\midrule
High School Microeconomics & Sonnet3.5 & 0.750 & -0.096 & 0.042 & -2.307 & \textbf{0.036} & \textbf{-0.512} & 0.262 & 0.213 \\
 & Aya23 35B & 0.549 & -0.105 & 0.038 & -2.775 & \textbf{0.014} & \textbf{-0.582} & 0.339 & 0.295 \\
 & DeepSeek R1 & 0.888 & -0.088 & 0.074 & -1.194 & 0.251 & -0.295 & 0.087 & 0.026 \\
 & DeepSeek V3 & 0.906 & -0.157 & 0.049 & -3.179 & \textbf{0.006*} & \textbf{-0.634} & 0.403 & 0.363 \\
 & Gemini 1.5 Pro & 0.883 & -0.129 & 0.067 & -1.920 & 0.074 & -0.444 & 0.197 & 0.144 \\
 & Llama3.1 405B & 0.953 & -0.185 & 0.050 & -3.691 & \textbf{0.002*} & \textbf{-0.690} & \textbf{0.476} & \textbf{0.441} \\
 & Phi4 & 0.858 & -0.184 & 0.053 & -3.479 & \textbf{0.003*} & \textbf{-0.668} & 0.447 & 0.410 \\
 & GPT-4o & 0.942 & -0.150 & 0.084 & -1.779 & 0.096 & -0.417 & 0.174 & 0.119 \\
 & Pixtral 12B & 0.622 & -0.105 & 0.033 & -3.179 & \textbf{0.006*} & \textbf{-0.635} & 0.403 & 0.363 \\
 & Qwen2.5 32B & 0.779 & -0.154 & 0.048 & -3.196 & \textbf{0.006*} & \textbf{-0.636} & 0.405 & 0.365 \\
\midrule
International Law & Sonnet3.5 & 0.645 & -0.040 & 0.028 & -1.426 & 0.174 & -0.346 & 0.119 & 0.061 \\
 & Aya23 35B & 0.578 & -0.101 & 0.042 & -2.403 & \textbf{0.030} & \textbf{-0.527} & 0.278 & 0.230 \\
 & DeepSeek R1 & 0.813 & -0.043 & 0.043 & -1.010 & 0.329 & -0.252 & 0.064 & 0.001 \\
 & DeepSeek V3 & 0.771 & -0.073 & 0.045 & -1.617 & 0.127 & -0.385 & 0.148 & 0.092 \\
 & Gemini 1.5 Pro & 0.796 & -0.039 & 0.052 & -0.758 & 0.460 & -0.192 & 0.037 & -0.027 \\
 & Llama3.1 405B & 0.876 & -0.096 & 0.038 & -2.548 & \textbf{0.022} & \textbf{-0.550} & 0.302 & 0.256 \\
 & Phi4 & 0.804 & -0.101 & 0.041 & -2.452 & \textbf{0.027} & \textbf{-0.535} & 0.286 & 0.238 \\
 & GPT-4o & 0.889 & -0.085 & 0.072 & -1.175 & 0.258 & -0.290 & 0.084 & 0.023 \\
 & Pixtral 12B & 0.686 & -0.095 & 0.033 & -2.859 & \textbf{0.012*} & \textbf{-0.594} & \textbf{0.353} & \textbf{0.310} \\
 & Qwen2.5 32B & 0.787 & -0.092 & 0.040 & -2.297 & \textbf{0.036} & \textbf{-0.510} & 0.260 & 0.211 \\
\bottomrule
\end{tabular}
}
\caption{Fertility and Accuracy by Model and Subject. Results from linear models regressing accuracy on fertility across 16 languages for each model-subject combination. The table reports intercepts, slopes (negative values indicate higher fertility correlates with lower accuracy), and standard errors, $t$-statistics, and $p$-values for the slopes in each regression. Pearson 
correlation coefficients ($\rho$) quantify the strength of the linear relationship. $R^2$ values show the proportion of variance in accuracy explained by fertility, and adjusted $R^2$ values correct for sample size. Bold $p$-values indicate statistical significance ($p < 0.05$). Asterisks (*) indicate results that remain significant after Benjamini-Hochberg FDR correction (FDR $< 0.05$). Bold $\rho$ values indicate strong correlations ($|\rho| \geq 0.50$), and bold $R^2$ values indicate large effects ($R^2 \geq 0.25$). Regressions for o1 are not included because OpenAI has not released details on the tokenizer for this model.}
\label{tab:fertility_accuracy_regression}
\end{table*}

\subsection{Performance Gaps Across Languages}

\begin{figure*}[!htbp]
    \centering
    \begin{subfigure}[t]{0.49\textwidth}
        \centering
        \includegraphics[width=\linewidth]{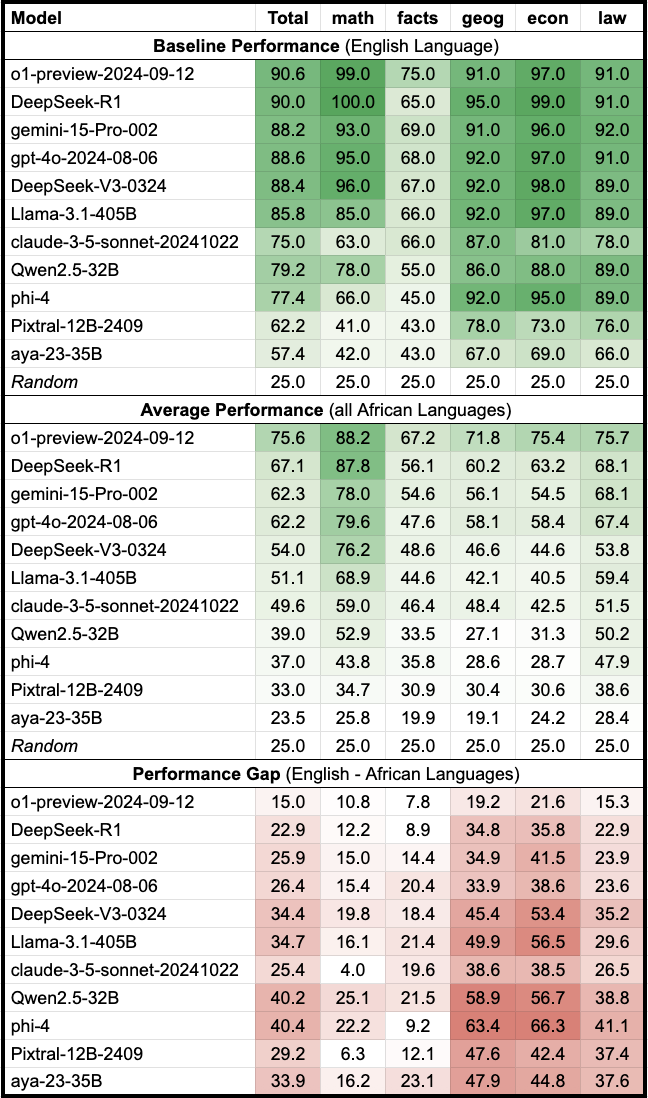}
        \caption{Accuracy Aggregation (English)}
        \label{fig:acc_agg_eng}
    \end{subfigure}%
    \hfill
    \begin{subfigure}[t]{0.49\textwidth}
        \centering
        \includegraphics[width=\linewidth]{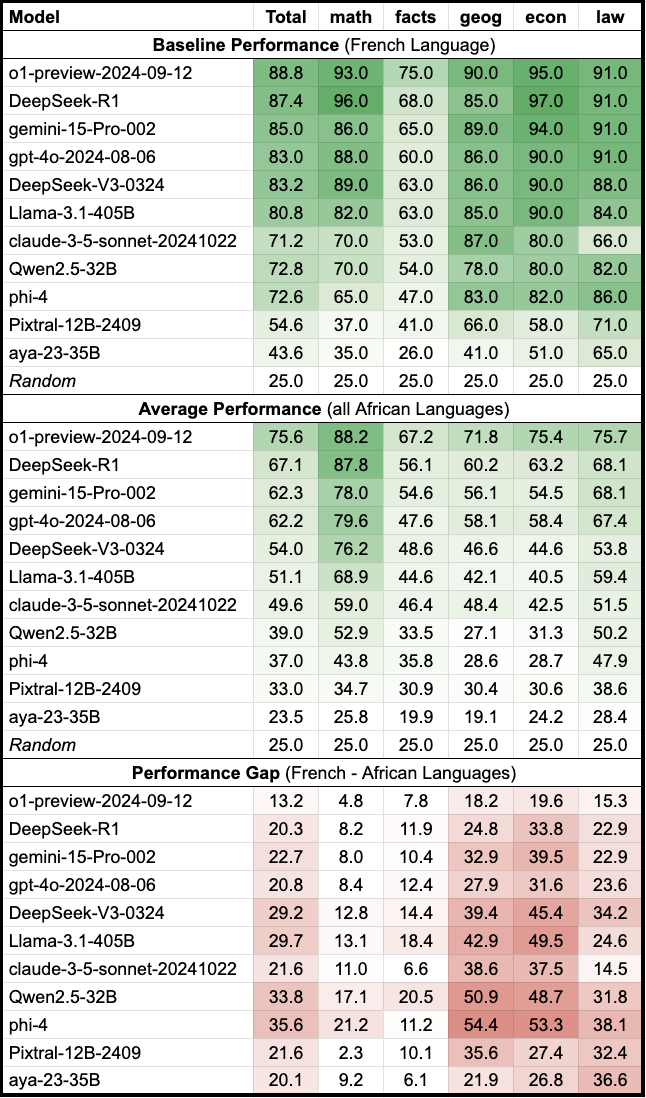}
        \caption{Accuracy Aggregation (French)}
        \label{fig:acc_agg_fre}
    \end{subfigure}
    \caption{Baseline performance shows English (a) and French (b) accuracy (in percentage points). The mean accuracy across all 16 African languages is shown in the middle charts of (a) and (b). The bottom charts of (a) and (b) show performance gaps between the African languages and higher-resource languages, though reasoning-oriented models narrow this gap.}
    \label{fig:accuracy_aggregation_comparison}
\end{figure*}

Figure \ref{fig:accuracy_aggregation_comparison} shows substantial performance disparities across languages. African languages trail English by 30 percentage points and French by 24 percentage points on average across model results. This gap varies by subject, with Geography and Economics showing the largest disparities and Elementary Mathematics and Global Facts showing relatively smaller gaps.

Reasoning models DeepSeek R1 and o1 reduce performance gaps across subjects on average.  These models have higher average accuracy than non-reasoning counterparts while maintaining strong English performance, suggesting that enhanced reasoning capabilities provide particular benefits for low-resource settings.

% \begin{figure}[!htbp]
%     \centering
%     \begin{subfigure}[t]{0.45\textwidth}
%         \centering
%         \includegraphics[width=\linewidth]{figures/acc_agg_eng.png}
%         \caption{Accuracy Aggregation (English)}
%         \label{fig:acc_agg_eng}
%     \end{subfigure}%
%     \hfill
%     \begin{subfigure}[t]{0.45\textwidth}
%         \centering
%         \includegraphics[width=\linewidth]{figures/acc_agg_fra.png}
%         \caption{Accuracy Aggregation (French)}
%         \label{fig:acc_agg_fre}
%     \end{subfigure}
%     \caption{Baseline performance shows English (a, top) and French (b, top) accuracy (in percentage points).  The mean accuracy across all 16 African languages is shown in the middle charts of (a) and (b).  The bottom charts of (a) and (b) show performance gaps between the African languages and higher-resource languages, \red{though reasoning-oriented models substantially narrow this gap}.}
%     \label{fig:accuracy_aggregation_comparison}
% \end{figure}

\begin{figure*}[!htbp]
\centering
\includegraphics[width=0.95\textwidth]{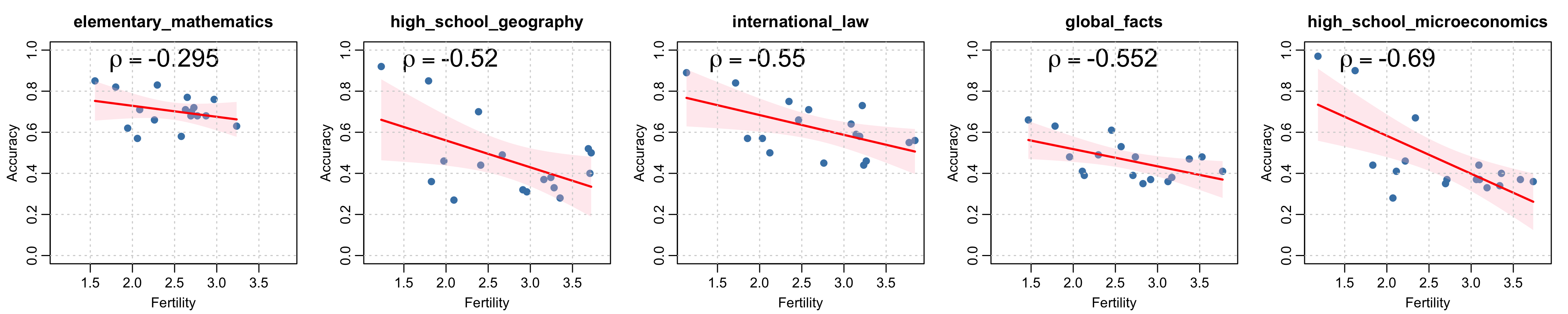}
\caption{Fertility and accuracy for Llama 3.1 405B across subjects. Correlations ($\rho$) are negative in all five subjects, from $-0.30$ (Elementary Math) to $-0.69$ (Microeconomics). Fertility is not independent of training data representation; fertility captures tokenization inefficiency that covaries with performance, but does not isolate causal effects independent of pretraining data availability or quality.}
\label{fig:llama31_405b_fertility_accuracy}
\end{figure*}

\subsection{Tokenization as Performance Predictor}
Figure \ref{fig:llama31_405b_fertility_accuracy} demonstrates the negative relationship between fertility and accuracy for Llama 3.1 405B as an example model (results for all models are displayed in Figure \ref{fig:all_models_combined_grid}). Across all 10 models and five subjects, the fertility--accuracy slope is negative in every regression (50 of 50), ranging from $-0.002$ to $-0.185$. The association is statistically significant in 24 of the 50 regressions at $p<0.05$; among the 12
regressions that survive FDR correction, slopes range from $-0.079$ to $-0.185$
(8 to 18.5 percentage points of accuracy per additional token per word).  The association is weak or absent in some cases: for DeepSeek R1 on Global Facts the slope is near zero ($-0.002$, $p=0.952$), and no Global Facts regression survives FDR correction (mean slope $-0.036$, versus $-0.135$ for Microeconomics).

We note that fertility is not independent of language resource level, that is languages underrepresented in training corpora tend to have higher fertility due to fewer subword merges during tokenizer training.  The observed fertility accuracy relationship is likely reflecting both the direct tokenization effects and indirect effects mediated by training data availability.

Table \ref{tab:fertility_accuracy_regression} reports detailed regression results. Several effects are both large and statistically significant after FDR correction, including Llama-3.1-405B on Microeconomics (slope $= -0.185$, $p = 0.002$) and Qwen-2.5-32B on Geography (slope $= -0.155$, $p = 0.006$). Where the association is significant (24 regressions), fertility explains 24–52\% of the variance in accuracy; across all 50 regressions R² ranges from 0.000 to 0.523, and 14 are below 0.10. Effects are largest in Geography and Microeconomics (mean slopes $-0.117$ and $-0.135$) and smallest in Global Facts ($-0.036$).

\section{Analysis}

\subsection{Economic Impact of Token Inflation}

The tokenization inefficiencies documented above translate directly to economic barriers. Because transformer training scales quadratically with sequence length, a 2× increase in fertility produces a 4× increase in training time and cost. Table \ref{tab:training} quantifies these impacts for the Llama model family.

Inference costs show similar patterns. Generating 1M English-equivalent tokens costs \$5-20 with GPT-4o, but \$10-40 for a language with twice the token fertility. Latency doubles correspondingly, creating user experience degradation alongside cost inflation.

These cost projections assume equivalent training objectives across languages.  In practice, the relationship between fertility and cost interacts with data availability: high-fertility languages lack training data to realize theoretical training costs, which is an additional and compounded barrier to development.  

\subsection{Linguistic Factors Driving Fertility}

The fertility disparities correlate strongly with morphological typology. Agglutinative languages in our dataset (Swahili: $F=2.8$, Zulu: $F=2.6$) consistently show higher fertility than more analytic languages. Fusional languages like Amharic occupy intermediate positions.  This pattern suggests that current BPE-based tokenizers, predominantly trained on English and European language text, systemically fail to capture morphological processes.  Affixation patterns that would constitute single semantic units are split across tokens, fragmenting the meaning and increasing the sequence length.

For example, Swahili verb forms that encode subject, tense, object, and mood in a single word are often split into multiple tokens, while equivalent English constructions using auxiliary verbs and pronouns may use fewer tokens despite containing more words. This fundamental mismatch between tokenizer training data and target language morphology drives the observed inefficiencies.

\section{Conclusion}

This study finds that higher tokenization inefficiency (fertility) is consistently associated with lower accuracy for morphologically complex, low-resource languages, although the association is statistically significant in about half of the model–subject regressions. Across 10 models and 16 African languages, fertility explains up to 52\% of the variance in accuracy in the strongest cases and little in others. In the strongest cases, each additional token per word is associated with accuracy drops of up to 18.5 percentage points.

While reasoning models like DeepSeek R1 and o1 narrow the English--African gap (15 and 23 points, versus 25--40 for the other nine models), substantial disparities remain. The economic implications are severe: doubling fertility quadruples training costs, creating a "token tax" that turns linguistic diversity into computational liability.

Addressing these inequities requires coordinated intervention: technical advances in morphologically-aware tokenization, economic reforms to pricing structures, and expanded multilingual evaluation infrastructure. Without such efforts, billions of speakers will remain excluded from the benefits of language technology, perpetuating digital divides along linguistic lines.

\section{Limitations}
Our analysis establishes correlation between fertility and accuracy but cannot establish causation. Token fertility correlates with training data availability, where languages with less representation in pretraining corpora receive fewer subword merges during tokenizer training, and have consequent higher fertility.  The performance degradation observed may reflect tokenizer inefficiency, scant data, or a combined effect. Also, while we report aggregate performance for such models where possible, fine-grained fertility analyses cannot be conducted without tokenizer transparency.

In terms of access to model components, we could not analyze OpenAI's o1 model tokenizer as implementation details remain proprietary, limiting our ability to fully characterize reasoning models' tokenization strategies.  Our analysis focuses on African languages which limits generalization to other language families.  It's possible that similar patterns would be present for morphologically complex, low-resource languages, this is untested.  While we analyze 16 African languages, this represents less than 1\% of Africa's 2,000+ languages. Our findings may not generalize to languages with different morphological properties or non-Latin scripts.  We evaluate only multiple-choice question answering, MCQA tasks. Performance degradation patterns may differ for generative tasks where token inflation affects both input and output, potentially compounding the effects we observe, but this remains speculation.  Our cost calculations use publicly available pricing and may not reflect negotiated rates or future hardware improvements. We focus on direct computational costs without quantifying broader environmental impacts.

By quantifying performance gaps, we risk reinforcing perceptions that some languages are lesser for NLP applications. We emphasize that these disparities reflect technological limitations, not inherent language properties.  Documenting the "token tax" could discourage investment in low-resource language technologies if stakeholders focus solely on costs rather than equity. This document is intended to motivate solutions.  To address these risks, we suggest developing inclusive tokenization standards through community participation; advocating for subsidized compute resources for low-resource language research; and creating evaluation metrics that explicitly penalize tokenization inefficiency.  

\bibliography{main}

\appendix

\section*{Appendix}

\section{Regression Results by Subject}

\begin{figure*}[!htbp]
\centering
\includegraphics[width=0.7\textwidth]{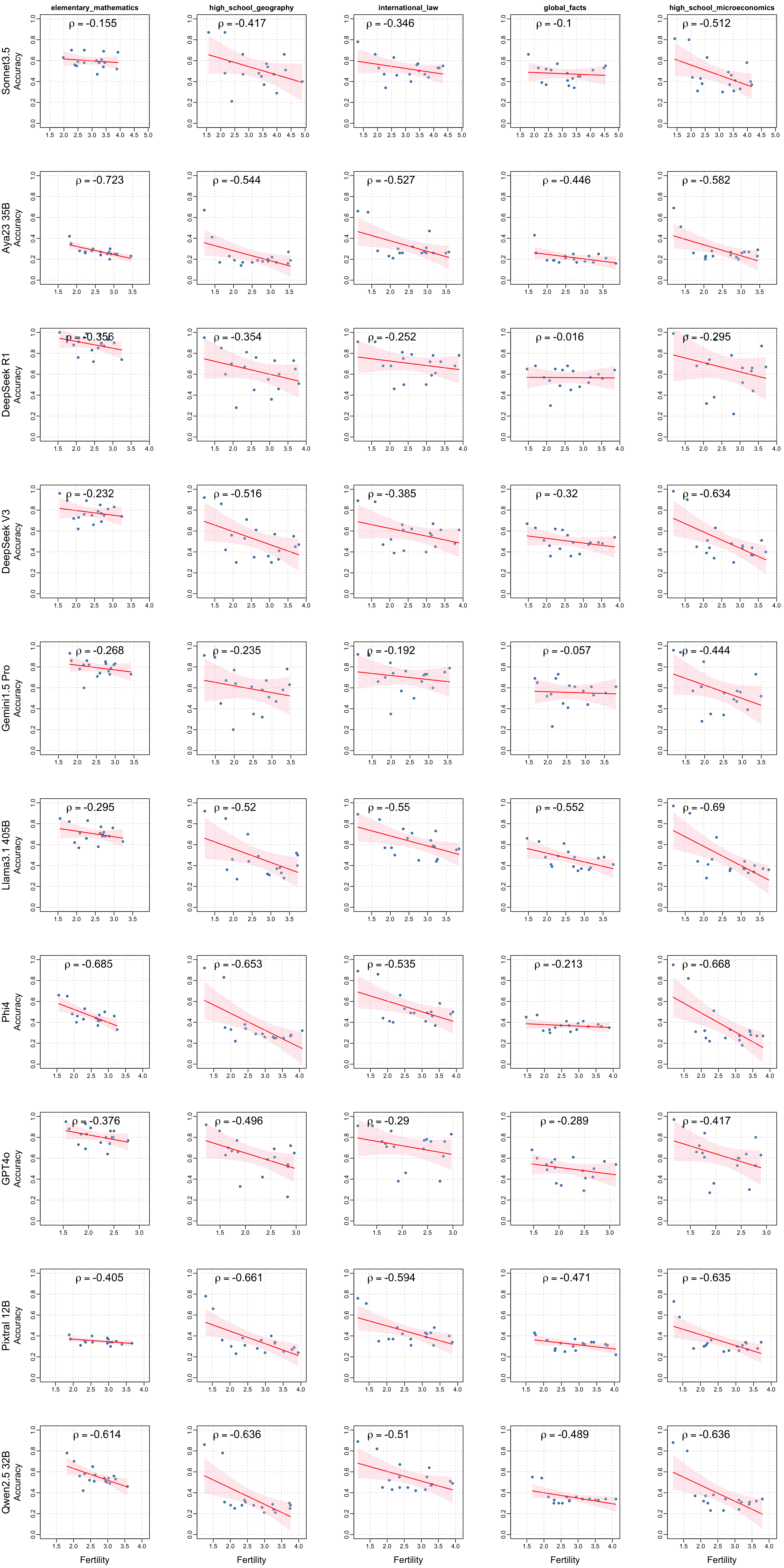}
\caption{Fertility and accuracy trade-offs for the 10 models across five MMLU subjects. Note that these associations are correlational: fertility captures tokenization inefficiency that covaries with performance, but does not isolate causal effects independent of pretraining data availability or quality.}
\label{fig:all_models_combined_grid}
\end{figure*}

Table \ref{tab:fertility_accuracy_regression}  and Figure \ref{fig:all_models_combined_grid}  show results of accuracy-on-fertility regressions for the 10 models over 5 subjects.

\section{Fertility}
Fertility measures the average number of tokens required to represent a word in a corpus:
\[
F = \frac{T}{W}
\]
for \(T\) and \(W\) the token and word counts.  Higher
\(F\) inflates sequence length, affecting a model's ability to learn long-range dependencies and compute
costs \citep{ali-etal-2024-tokenizer}.

\section{Inference}
\subsection{Training and Inference Cost Comparison: English vs.\ Language X}

This is a thought exercise in training and inference costs for LLMs applied to English and Language X. The analysis assumes the same model architecture and tokenizer across languages, with cost differences due to tokenization inefficiencies and quadratic $O(n^2)$ training scaling \cite{vaswani2023attentionneed} of transformer models. 

We assume Language X has a fixed \textbf{2× increase} in tokens across tokenizers (although there are variations not included here).  
We assume English has 1\,000\,000 tokens (baseline) and Language X: approximately 2\,000\,000 tokens for equivalent content.  There is a $2^2 = 4$x increase in training cost.

\begin{table*}[!htbp]
\centering

\begin{tabular}{lccc}
\toprule
\textbf{Model} & \textbf{English} & \textbf{2× Fertility} & \textbf{5× Fertility} \\
\midrule
Llama 2 70B & \$5M & \$20M & \$125M \\
Llama 3 70B & \$24M & \$96M & \$600M \\
Llama 3.1 405B & \$105M & \$420M & \$2.6B \\
\bottomrule
\end{tabular}
\caption{Training costs scale quadratically with fertility}
\label{tab:training}
\end{table*}

% \subsection{Training Compute Cost (Open-Weight LLaMA Models)}

% Using published petaFLOP-day figures and assuming a compute cost of \$240 per petaFLOP-day:

\begin{table*}[!htbp]
\centering
\begin{tabular}{lccc}
\hline
\textbf{Model} & petaFLOP-days & English \$ & Language X (\$4$\times$) \\
\hline
LLaMA 2 (69B)     & 21\,000   & 5 M     & 20 M     \\
LLaMA 3 (70B)     & 100\,000  & 24 M    & 96 M     \\
LLaMA 3.1 (405B)  & 440\,000  & 105 M   & 420 M     \\
\hline
\end{tabular}
\caption{Training compute and cost estimates for LLaMA models (USD).}
\label{tab:training2}
\end{table*}

\begin{table*}[!htbp]
\centering
\small

%\resizebox{\textwidth}{!}{%
\begin{tabular}{@{} l l l l @{}}
\toprule
Provider & Model (type) & English \$ & Language~X~ ($\sim$2$\times$) \\
\midrule
OpenAI & \makecell[tl]{GPT-4o} & 5 / 20 & 10 / 40 \\
OpenAI & \makecell[tl]{o4-mini*} & 4 / 16 & 8 / 32 \\
Google & \makecell[tl]{Gemini 2.5 Flash} & 0.30 / 2.50 & 0.60 / 5.00 \\
Google & \makecell[tl]{Gemini 2.5 Pro*} & 1.25 / 10 & 2.50 / 20 \\
Anthropic & \makecell[tl]{Claude 4 Sonnet} & 3 / 15 & 6 / 30 \\
Anthropic & \makecell[tl]{Claude 4 Opus*} & 15 / 75 & 30 / 150 \\
\bottomrule
\end{tabular}%} 

\caption{Inference cost per 1M English-equivalent tokens (USD) including reasoning models. The costs are shown for input/output.}
\label{tab:inference-reasoning}
\end{table*}

In addition to cost, token inflation impacts time.  Transformer models scale quadratically in sequence length.  With a 2$\times$ token increase, Language~X requires $4\times$ more compute. This means training that takes 90 days for English would take $\sim$360 days for Language~X on the same hardware.
For inference time, decoding scales approximately linearly with token count.  A prompt completion that takes 2 seconds in English may take about 4 seconds in Language~X.

These multipliers apply whether the additional tokens appear in the input (prompt) or output (completion), and they exacerbate cost disparities for low-resource languages.

Using published petaFLOP-day figures and assuming a compute cost of \$240 per petaFLOP-day, Table~\ref{tab:training2} displays order-of-magnitude estimates of compute and cost for training, while Table~\ref{tab:inference-reasoning} displays order-of-magnitude estimates for inference.

\subsection{Prompt}
\begin{small}
\begin{verbatim}
You must only reply with 'Final Answer: X' 
where X is A, B, C, or D.
Do NOT add explanations, reasoning, 
or extra text.
Question: <question text>
Choices:
A. <option 1>
B. <option 2>
C. <option 3>
D. <option 4>
Your response must be strictly formatted as:
Final Answer: X
\end{verbatim}
\end{small}

\section{Reproducibility: Code and Data}

Raw LLM outputs, tokens, fertility and parity measures for each model, and scripts to replicate our analysis are available at:\\ \url{https://github.com/jessicalundin/multilingual_token_tax}.

\end{document}